\documentclass[sigconf,nonacm]{acmart}

\usepackage{booktabs}
\usepackage{enumitem}
\usepackage[most]{tcolorbox}
\usepackage{tikz}
\usetikzlibrary{arrows.meta, positioning, shapes.geometric, fit, backgrounds}
\usepackage{balance}
\usepackage{xspace}
\usepackage{pifont}

\newcommand{\dataset}{\textsc{GF-Bench}\xspace}
\newcommand{\datasetfullname}{\textbf{G}PU \textbf{F}ailure \textbf{Bench}mark}

\newcommand{\mathbbm}[1]{\text{\usefont{U}{bbm}{m}{n}#1}} 

\makeatletter
\renewcommand{\bibliofont}{\scriptsize}
\makeatother

\begin{document}

\title{When Log Parsing Removes What Matters: Detecting Failures in Distributed GPU Workloads}

\author{Boyuan Cui}
\orcid{0009-0008-9367-8457}
\affiliation{%
  \institution{University of Toronto}
  \city{Toronto}
  \country{Canada}
}
\email{boyuanbryan.cui@mail.utoronto.ca}

\author{Yuyang Zeng}
\orcid{0009-0008-3291-0507}
\affiliation{%
  \institution{University of Toronto}
  \city{Toronto}
  \country{Canada}
}
\email{ryanyy.zeng@mail.utoronto.ca}

\author{Yifang Tian}
\authornote{Corresponding author.}
\orcid{0000-0002-8253-2064}
\affiliation{%
  \institution{University of Toronto}
  \city{Toronto}
  \country{Canada}
}
\email{yifang.tian@mail.utoronto.ca}

\author{Hans-Arno Jacobsen}
\orcid{0000-0003-0813-0101}
\affiliation{%
  \institution{University of Toronto}
  \city{Toronto}
  \country{Canada}
}
\email{jacobsen@eecg.toronto.edu}

\renewcommand{\shortauthors}{Cui, Zeng, Tian, Jacobsen}

\begin{abstract}
Large AI models increasingly rely on distributed GPU execution, yet it remains
unclear whether established anomaly detection methods transfer effectively to this
setting. We introduce \dataset, a PyTorch-based framework for generating controlled 
failures in distributed GPU workloads, and a dataset produced with it. The framework
supports extensible workload and fault-injection logic; based on this design, the
current release targets failures visible in multi-GPU communication logs and provides 220 labeled two-GPU
runs covering normal behavior, CUDA out-of-memory fail-stop faults, and
persistent-straggler fail-slow faults. Using \dataset, we evaluate three
sequence-based methods (DeepLog, LogAnomaly, and LogBERT) and the large language model (LLM)-based
LogPrompt with in-context learning (ICL) and chain-of-thought (CoT) prompting strategies on parsed logs, and additionally evaluate LogPrompt on raw logs.
On parsed inputs, no method reliably detects both failure
classes, and the highest overall F1 score is 0.667. Raw inputs raise the overall
F1 score of ICL-LogPrompt to 0.983, at $9.0\times$ its average input-token
usage and $8.9\times$ its per-run cost. These results 
reveal a trade-off between effectiveness and efficiency: 
parsed inputs reduce LLM token use and inference cost
but can discard the distributed execution context required for
accurate failure detection.
\dataset provides a reproducible basis for studying
cost-efficient anomaly detection methods and can be extended to additional GPU workloads and failure types.

\end{abstract}

\keywords{distributed GPU system, NCCL, log-anomaly detection, benchmark, fault injection, failure analysis}

\ccsdesc[500]{Software and its engineering~Software maintenance tools}
\ccsdesc[500]{Computer systems organization~Dependable and fault-tolerant systems and networks}
\maketitle

\begingroup
\renewcommand{\thefootnote}{}
\footnotetext{%
An earlier version of this paper appeared in
\textit{Proceedings of the 1st International Workshop on
Reliable and Trustworthy Automated Software Engineering
(RASE '26)}.
\url{https://doi.org/10.1145/3820756.3844919}
}
\endgroup

\section{Introduction}
Distributed training is a core component of modern AI infrastructure. Large training campaigns can occupy thousands of GPUs for extended periods, so a single failure or persistent slowdown can waste substantial computation and delay completion~\cite{llama3herdmodels}. Communication failures are particularly important in distributed settings because workers synchronize repeatedly, making GPU communication central to both performance and reliability~\cite{yu2024survey}. 
On NVIDIA-based clusters, communication operations are commonly implemented through the NVIDIA Collective Communications Library (NCCL)~\cite{nccl}. NCCL logs record communicator initialization, collective progress, and shutdown behavior, providing useful evidence for failure analysis and detection.

Prior work has developed mature techniques and benchmarks for log-based anomaly detection and related AI for IT operations (AIOps) tasks. DeepLog~\cite{deeplog}, LogAnomaly~\cite{loganomaly}, and LogBERT~\cite{logbert} learn normal event-sequence patterns from parsed templates. LogPrompt~\cite{liu2024logprompt} applies prompt-based classification with a large language model (LLM). Loghub~\cite{loghub} provides a broad corpus for system-log analytics, RCAEval~\cite{pham2025rcaeval} benchmarks telemetry-based root-cause analysis in microservices, and AIOpsLab~\cite{chen2025aiopslab} provides reproducible cloud environments, fault injection, workloads, and telemetry for evaluating AIOps agents. However, these methods and benchmarks primarily target CPU-based systems rather than distributed GPU workloads.

The move toward LLM-based approaches also introduces an operational constraint that earlier methods did not face. Querying an LLM remains substantially more expensive than running a trained sequence model, a cost that is difficult to sustain for systems that emit tens of gigabytes of logs per hour~\cite{jiang2024lilac}. Existing systems reduce LLM overhead in different ways. LILAC~\cite{jiang2024lilac} uses an adaptive parsing cache to reuse previously generated and refined log templates, reducing redundant LLM queries, while Tabi~\cite{wang2023tabi_cost1} uses a multi-level inference engine that serves most queries with smaller models and reroutes low-confidence cases to an LLM. Both systems demonstrate substantial efficiency gains, but how log representation affects the cost-effectiveness of LLM-based failure detection remains underexplored for distributed GPU logs.

A separate line of GPU reliability research provides hardware-oriented fault-injection tools: SASSIFI~\cite{hari2017sassifi} instruments low-level GPU assembly to inject errors into architecture-visible state, while NVBitFI~\cite{tsai2021nvbitfi} dynamically instruments GPU binaries to inject transient faults. These tools, however, focus on low-level fault injection and lack an end-to-end workflow spanning multi-GPU workloads, fault injection, observability collection, and anomaly detection.

Existing work therefore offers either general-purpose log-anomaly detection techniques or low-level GPU fault injection, with no common ground between them: no framework produces controlled, labeled failures in distributed GPU workloads, and consequently no basis exists for measuring how well existing analysis techniques detect such failures or what those techniques cost to run. 

We address this with \dataset (\datasetfullname), a framework built on PyTorch~\cite{paszke2019pytorch} that executes multi-GPU workloads, injects faults at controlled points, and collects the resulting observability data. \dataset separates workload definition, fault injection, and log collection so that each can be extended independently. 
The initially released dataset with NCCL-visible failures covers two fault types that instantiate the fail-stop and fail-slow modes distinguished in prior reliability work~\cite{gunawi2018failslow}: CUDA out-of-memory and a persistent straggler. Using this release, we evaluate four log-anomaly detection methods on logs parsed with Drain~\cite{drain} and additionally evaluate LogPrompt on raw logs. Our contributions are:
\begin{enumerate}[leftmargin=*]
  \item \textbf{Framework.} \dataset, a PyTorch-based framework that executes configurable multi-GPU workloads and automates fault injection and log collection.
  \item \textbf{Dataset.} A dataset of 220 labeled runs produced by \dataset, covering normal behavior, CUDA out-of-memory (OOM) fail-stop faults, and persistent-straggler fail-slow faults.
  \item \textbf{Detection Effectiveness.} A comparison of four log-anomaly detection methods under a common parsed-log protocol.
  \item \textbf{Cost of Log Representation.} A measurement of what changes when parsing is removed, quantifying the detection effectiveness gained and the token and monetary cost incurred.
\end{enumerate}

\section{\dataset Design}
\noindent\textbf{Scope and task definition.}
For a multi-GPU run $i$ executed on $W_i$ GPUs, let $L_{i,w}$ denote the observability data (logs, metrics, traces) collected from GPU $w\in\{0,\ldots,W_{i} - 1\}$ and its associated process. Let $\mathcal{L}_i=\{L_{i,0},\ldots,L_{i,W_{i} - 1}\}$ denote all the observability data produced by run $i$, and let $y_i\in\mathcal{Y}=\{0,1,\ldots,K\}$ denote the label of the run, where $y_i=0$ denotes normal execution and $y_i\in \{1,\ldots,K\}$ denotes one of $K$ failure classes such as CUDA OOM or a persistent straggler. We write $a_i=\mathbbm{1}[y_i\geq 1]$ for the induced binary anomaly label.
A representation strategy $s\in\mathcal{S}$, such as log parsing, transforms the observability data into a detector input $x_i^{(s)}=s(\mathcal{L}_i)$, and a detector $f\in\mathcal{F}$ predicts $\hat{a}_i=f(x_i^{(s)})\in\{0,1\}$.

\dataset generates labeled failure data so that each detector and representation
pairing $(f,s)$ can be assessed on both predictive quality and 
operational cost. The objective is to characterize the trade-off
between the two.

\noindent\textbf{Execution framework.}
We build \dataset (Figure~\ref{fig:workflow}), a PyTorch-based framework for executing multi-GPU runs with injected faults. By leveraging PyTorch's device abstraction, our framework adapts to different GPU environments with minimal changes to the core implementation.

In our framework, each run launches a supervising driver process that spawns one process per GPU, where each process is bound to a dedicated GPU and assigned a rank: a zero-based integer identifier used by NCCL to coordinate collective communication across processes~\cite{nvidia_nccl_usage_rank}. The framework supports customizable GPU workloads with varied computation and communication patterns. For example, to simulate distributed AI training, the workload can execute multiple iterations, each combining tensor computations that approximate neural network operations (e.g., forward/backward passes) with collective communication (e.g., AllReduce) representing gradient and parameter synchronization.

Each GPU workload exposes a fault-injection interface with two components: a condition checker that determines when to trigger a fault, and an injection function that implements the fault behavior.
The driver process for each run also enforces a customizable global timeout. If this timeout is reached before all GPU workloads complete, the driver forcibly terminates all spawned processes. This guards against indefinite hangs that faults such as stragglers or crashed collectives can otherwise induce, and ensures every run terminates and is recorded with a definite outcome.

\newcommand{\diagramscale}{0.7} 
\begin{figure}[h]
\resizebox{\diagramscale\width}{!}{%
\begin{tikzpicture}[
    font=\sffamily\normalsize,
    box/.style={
        draw, thick, rounded corners=2pt,
        align=center, minimum height=0.8cm, inner sep=4pt
    },
    driver/.style={box, fill=blue!8, minimum width=5.6cm},
    workerhdr/.style={box, fill=orange!15, minimum width=4.4cm, minimum height=0.65cm},
    stepbox/.style={box, fill=white, minimum width=4.0cm, minimum height=0.6cm},
    hookbox/.style={draw, thick, dash pattern=on 3pt off 2pt, rounded corners=3pt,
        red!55!black, fill=red!3},
    hookstep/.style={box, fill=white, minimum width=2.6cm, minimum height=0.55cm},
    noopstep/.style={box, fill=white, minimum width=1.6cm, minimum height=0.5cm},
    arr/.style={-Stealth, thick},
]
 
\node[driver] (driver) {Supervising Driver Process\\\small enforces global timeout};
 
\node[minimum width=5.4cm, minimum height=2.1cm, below=0.50cm of driver] (cardgeom) {};
 
\draw[rounded corners=3pt, thick, draw=black!55, fill=orange!4]
    ([xshift=0.24cm,yshift=-0.24cm]cardgeom.north west)
    rectangle ([xshift=0.24cm,yshift=-0.24cm]cardgeom.south east);
\draw[rounded corners=3pt, thick, draw=black!55, fill=orange!8]
    ([xshift=0.12cm,yshift=-0.12cm]cardgeom.north west)
    rectangle ([xshift=0.12cm,yshift=-0.12cm]cardgeom.south east);
\draw[rounded corners=3pt, thick, draw=black!55, fill=orange!15]
    (cardgeom.north west) rectangle (cardgeom.south east);
 
\node[workerhdr, below=0.7cm of driver] (whdr) {Worker Process};
 
\node[stepbox, below=0.3cm of whdr] (comp) {GPU Workload};
 
\draw[arr] (whdr.south) -- (comp.north);
 
\draw[arr] (driver.south) -- (whdr.north);
\path (driver.south) -- (whdr.north) coordinate[pos=0.35] (spawnpos);
\node[fill=white, inner sep=1.5pt, font=\small, anchor=west, xshift=1pt] at (spawnpos) {spawn};
 
\node[diamond, draw=red!55!black, thick, fill=white, aspect=2.4,
      align=center, font=\small, inner sep=2pt,
      right=1.6cm of cardgeom, anchor=west, yshift=0.15cm] (condc) {Condition\\Checker};
 
\node[hookstep, draw=red!55!black, below=0.5cm of condc] (inj) {Injection Function};
\node[noopstep, right=0.5cm of condc] (noop) {No-op};
 
\draw[-Stealth, thick, red!55!black] (condc.south) -- node[left, xshift=-0.03cm]{\small yes} (inj.north);
\draw[arr] (condc.east) -- node[above, font=\small]{no} (noop.west);
 
\begin{pgfonlayer}{background}
\node[hookbox, fit=(condc)(inj)(noop), inner xsep=10pt, inner ysep=10pt] (hookfit) {};
\end{pgfonlayer}
 
\node[anchor=south, font=\small\bfseries, red!55!black, yshift=2pt] (hooktitle)
    at (hookfit.north) {Fault Injection Interface};
 
\node[circle, fill=black, inner sep=0pt, minimum size=2.2pt] (hookdot) at (comp.east) {};
\draw[thick, red!55!black] ([yshift=0.15cm]hookdot.center) arc (90:370:0.085);
\draw[red!55!black, thick, dash pattern=on 3pt off 2pt]
    (hookdot.center) -- (hookfit.west);
 
\end{tikzpicture}
}
    \vspace{-2mm}
\caption{\dataset execution framework.}
\label{fig:workflow}
\end{figure}

\noindent\textbf{Data collection pipeline.}
On top of the \dataset execution framework, we implement a data collection pipeline to capture observability data generated from each run. GPU workloads typically expose telemetry through multiple channels, including standard output streams and dedicated log files. Our pipeline automatically captures and redirects these outputs to a designated directory for each experiment run. For example, NCCL provides detailed logs of collective communication behavior, which can be collected by configuring the NCCL\_DEBUG\_FILE environment variable.

The pipeline also supports systematic generation of diverse observability data by instantiating experiments with different workload and fault configurations. New workload behaviors and failure scenarios can be incorporated by implementing custom workload logic and a fault-injection interface. Furthermore, variations of the same fault type can be generated by modifying only the condition checker while reusing the same injection function. For example, an OOM fault can be triggered at different stages of training by changing the iteration at which the fault condition is satisfied.

\section{Evaluation and Results}
This study is guided by the following research questions:
\begin{enumerate}[leftmargin=*,label=\textbf{RQ\arabic*:}]
  \item How effective are established log-anomaly detection methods on Drain-parsed NCCL logs?
  \item How does Drain-based log parsing affect LogPrompt's detection effectiveness on NCCL logs?
  \item How does Drain-based log parsing affect the cost and latency of LLM inference?
\end{enumerate}

\subsection{Experimental Setup}
\label{subsec:setup}
\noindent\textbf{Workload and faults.} We instantiate the framework with a workload that
simulates data-parallel training: each iteration performs lightweight tensor
computations and a fixed-duration delay, followed by an AllReduce representing
gradient synchronization. We inject two fault types that can be deterministically reproduced in our multi-GPU environment and are representative of fail-stop and fail-slow behavior in distributed training: CUDA OOM and straggler faults, giving $K=2$ failure
classes. A CUDA OOM fault is parameterized by fault rank and
fault iteration; when the designated rank reaches that iteration, the injection
function repeatedly allocates tensors until allocation fails, producing
fail-stop behavior. A straggler fault is parameterized by fault rank and delay duration; the
designated rank adds a fixed delay before every subsequent collective, producing
persistent fail-slow behavior. We additionally include a normal scenario without any fault injection.

\noindent\textbf{Dataset.} We sweep four configuration parameters (total workload
iterations, fault rank, OOM injection iteration, and straggler delay), repeating
each configuration ten times as a pragmatic compromise between capturing run-to-run variation and limiting cloud cost. Table
\ref{tab:dataset} summarizes the dataset composition. The dataset is constructed on two
NVIDIA A40 GPUs on the Runpod cloud platform with CUDA 12.4, NCCL 2.20.5, and PyTorch 2.4.1, using
\path{NCCL_DEBUG=TRACE}, \path{NCCL_DEBUG_SUBSYS=ALL}, and
\path{NCCL_DEBUG_FILE=nccl_logs_%h_%p.txt}
to obtain detailed per-rank logs in separate files.

\begin{table}[h]
\caption{\dataset data composition.}
    \vspace{-3mm}
\label{tab:dataset}
\scriptsize
\setlength{\tabcolsep}{4.0pt}
\renewcommand{\arraystretch}{1.0}
\begin{tabular}{@{}lrl@{}}
\toprule
\textbf{Class} & \textbf{Runs} & \textbf{Varied factors} \\
\midrule
Normal & 100 & 10--100 in steps of 10 $\times$ 10 trials \\
CUDA OOM & 60 & 2 ranks $\times$ 3 injection iterations $\times$ 10 trials \\
GPU Straggler & 60 & 2 ranks $\times$ 3 delay durations $\times$ 10 trials \\
\bottomrule
\end{tabular}
\end{table}

\noindent\textbf{Methods and representations.} Here we evaluate DeepLog~\cite{deeplog}, LogAnomaly~\cite{loganomaly}, and LogBERT~\cite{logbert} on Drain-parsed logs, and LogPrompt~\cite{liu2024logprompt} on both Drain-parsed and raw logs. We run LogPrompt using GPT-5.4~\cite{openai_gpt54_2026} with both the chain-of-thought (CoT) and in-context learning (ICL) strategies from the original paper. For CoT, the LLM receives the log and a prompt with anomaly detection rules and outputs a binary normal/anomalous prediction with a short explanation. For ICL, it receives the log and a prompt containing labeled examples and outputs only a binary prediction. The LLM API calls use temperature 0.0, while \texttt{top\_p}, seed, and reasoning effort are left at their API defaults. We use snapshot \texttt{gpt-5.4-2026-03-05} for our experiments. Although the framework admits broader modalities, this study uses NCCL logs as the sole observability source.

\noindent\textbf{Protocol.} Half the dataset (50 normal, 30 OOM, and 30 straggler runs)
is held out for evaluation. The other half supplies training and prompt
material: its 50 normal runs train DeepLog, LogAnomaly, and LogBERT, which model
normal sequences only, and three of its runs, one per class, serve as ICL-LogPrompt
demonstrations. Because every method consumes a single sequence per run  and NCCL logs lack timestamps, per-rank logs
are concatenated in ascending rank order to provide a deterministic ordering. We report precision, recall, F1, and false positive rate (FPR)
with abnormal runs as the positive class. We also measure these metrics per failure type
to expose class-specific blind spots. Cost is measured as input/output token
counts, LLM latency, and monetary cost per run.

\subsection{Results}
\noindent\textbf{RQ1: Baseline Effectiveness.}
Table~\ref{tab:overall-results} reports results for all baselines with Drain-parsed inputs. ICL-LogPrompt obtains the highest overall F1 score (0.667), but it detects every straggler while missing every OOM run. CoT-LogPrompt exhibits the complementary bias: it attains 0.933 recall on OOM but detects no stragglers. LogBERT detects only one OOM run, resulting in 0.033 overall F1, while DeepLog and LogAnomaly predict every test run as normal. 
The stronger performance of ICL-LogPrompt likely stems from its use of one labeled example per class, whereas the sequence models are trained only on normal runs and CoT-LogPrompt receives no examples. This difference should be considered when comparing their results.
Nevertheless, no baseline reliably covers both fault classes under this Drain-parsed protocol.

\begin{table}[h]
\caption{Baseline performance overall and by fault type (P=Precision, R=Recall, FPR=False Positive Rate). 
The highest F1 score in each category is highlighted in bold.}
    \vspace{-3mm}
\label{tab:overall-results}
\scriptsize
\setlength{\tabcolsep}{7pt}
\renewcommand{\arraystretch}{1.0}
\begin{tabular}{@{}lccc@{}}
\toprule
\textbf{Baseline} & \textbf{Overall} & \textbf{CUDA OOM} & \textbf{GPU Straggler} \\
 & \scriptsize{(P / R / F1 / FPR)} & \scriptsize{(P / R / F1 / FPR)} & \scriptsize{(P / R / F1 / FPR)} \\
\midrule
LogBERT           & 1.000 / .017 / .033 / .00 & 1.000 / .033 / .065 / .00 & .000 / .000 / .000 / .00 \\
DeepLog           & .000 / .000 / .000 / .00 & .000 / .000 / .000 / .00 & .000 / .000 / .000 / .00 \\
LogAnomaly        & .000 / .000 / .000 / .00 & .000 / .000 / .000 / .00 & .000 / .000 / .000 / .00 \\
LogPrompt &  &  &  \\
\scriptsize{ - ICL}
    & 1.000 / .500 / \textbf{.667} / .00
    & .000 / .000 / .000 / .00
    & 1.000 / 1.000 / \textbf{1.000} / .00 \\
\scriptsize{ - CoT}
    & .378 / .467 / .418 / .92
    & .378 / .933 / \textbf{.538} / .92
    & .000 / .000 / .000 / .92 \\
\bottomrule
\end{tabular}
\end{table}

\noindent\textbf{RQ2: Effect of Drain-Based Parsing.}
Table~\ref{tab:logprompt-rawvsparsed-results} compares LogPrompt on parsed and raw inputs; raw logs raise overall F1 for both strategies. ICL shows the largest change, with overall F1 rising from 0.667 to 0.983 and OOM/straggler F1 both reaching 0.983. CoT improves from 0.418 to 0.547 overall and begins detecting stragglers, though its FPR rises to 0.98.

\begin{table}[]
\caption{LogPrompt performance: raw vs.\ parsed logs. The highest F1 score in each category is highlighted in bold.}
    \vspace{-3mm}
\label{tab:logprompt-rawvsparsed-results}
\scriptsize
\setlength{\tabcolsep}{8pt}
\renewcommand{\arraystretch}{1.0}
\begin{tabular}{@{}lccc@{}}
\toprule
\textbf{Variation} & \textbf{Overall} & \textbf{CUDA OOM} & \textbf{GPU Straggler} \\
\scriptsize{(Log type)} & \scriptsize{(P / R / F1 / FPR)} & \scriptsize{(P / R / F1 / FPR)} & \scriptsize{(P / R / F1 / FPR)} \\
\midrule
CoT \scriptsize{(Parsed)} & .378 / .467 / .418 / .92  & .378 / .933 / .538 / .92 & .000 / .000 / .000 / .92 \\
CoT \scriptsize{(Raw)}    & .456 / .683 / .547 / .98  & .372 / .967 / .537 / .98 & .197 / .400 / .264 / .98 \\
\cmidrule(lr){1-4}
ICL \scriptsize{(Parsed)}  & 1.000 / .500 / .667 / .00 & .000 / .000 / .000 / .00 & 1.000 / 1.000 / \textbf{1.000} / .00\\
ICL \scriptsize{(Raw)} & 1.000 / .967 / \textbf{.983} / .00 & 1.000 / .967 / \textbf{.983} / .00 & 1.000 / .967 / .983 / .00 \\
\bottomrule
\end{tabular}
\end{table}

To examine the reason behind the performance difference, we select one representative OOM run and ask the LLM to give a short reason along with the answer under the two input representations.
Figure~\ref{fig:icl-llm-answer-parsedvsraw} shows that with raw input, the model observes that rank 1 lacks the \texttt{Abort COMPLETE} lines and therefore identifies the execution as incomplete. With parsed input, the model fails to recognize this pattern.

In the collected logs, each message includes a host/process/thread prefix and an explicit source-rank field (e.g., \texttt{<host>:<pid>:<tid> [<rank>] NCCL ...}). In this case, Drain treats several of these fields as variables. As illustrated in Figure~\ref{fig:drain-template}, the two \texttt{Abort COMPLETE} messages from different ranks are mapped to the same template.

Source-rank identity is necessary for the evaluated OOM signature. This example indicates that Drain-based parsing can remove relational structure needed by downstream detection of distributed-training failures. This motivates parsing designed for GPU failures, preserving rank identity, collective identity, and alignment across per-rank logs.

\begin{figure}[]
    \begin{tcbraster}[raster columns=2, raster equal height,
                      raster left skip=1.5mm, raster right skip=1.5mm,
                      fontupper=\scriptsize, fonttitle=\scriptsize\bfseries,
                      boxsep=1pt, left=1pt, right=1pt, top=2pt, bottom=2pt]
    \begin{tcolorbox}[title=With Log Parser]
    This run matches the normal example almost exactly, including the full NCCL init, AllReduce, service thread closure, proxy exit, and \texttt{Abort COMPLETE} sequence, followed by the rank \texttt{[1]} informational lines. The same template set also appears in the provided normal example, and there are no extra error/failure patterns beyond the known benign plugin-missing message.\\
    \textcolor{red!70!black}{\textbf{\ding{55} Incorrect: Normal}}
    \end{tcolorbox}
    \begin{tcolorbox}[title=Without Log Parser]
    This run is semantically closer to the abnormal examples. Both ranks initialize successfully and begin AllReduce operations, but the log ends while rank 1 is still mid-sequence (\texttt{opCount b}) without the normal shutdown/abort-complete lines for rank 1. This truncated/incomplete execution pattern matches the abnormal examples more than the fully completed normal run.\\
    \textcolor{green!50!black}{\textbf{\ding{51} Correct: Anomalous}}
    \end{tcolorbox}
    \end{tcbraster}
        \vspace{-4mm}

    \caption{ICL-LogPrompt's answer for an OOM run with and without a Drain-based log parser.}
    \label{fig:icl-llm-answer-parsedvsraw}
\end{figure}

\begin{figure}[]
\begin{tcbraster}[raster columns=1, raster equal height, raster row skip=2pt,
    colframe=black, colback=white, boxrule=0.5pt, arc=1.0mm,
    fonttitle=\sffamily\bfseries\scriptsize, coltitle=black,
    colbacktitle=gray!15, top=1pt, bottom=1pt,
    fontupper=\ttfamily\scriptsize]
\begin{tcolorbox}[title=Log 1]
1eb576079025:19327:19511 [0] NCCL INFO comm 0x3c205590 rank 0 nranks 2 cudaDev 0 busId 4f000 - Abort COMPLETE
\end{tcolorbox}
\begin{tcolorbox}[title=Log 2]
1eb576079025:19328:19510 [1] NCCL INFO comm 0x48852150 rank 1 nranks 2 cudaDev 1 busId d1000 - Abort COMPLETE
\end{tcolorbox}
\end{tcbraster}
$\Downarrow$\ \textsf{\small Drain-based Log Parser}
\begin{tcolorbox}[colframe=black, colback=white, boxrule=1pt, arc=1.5mm,
    title=Extracted template, fonttitle=\sffamily\bfseries\scriptsize,
    coltitle=black, colbacktitle=gray!15, top=1pt, bottom=1pt,
    fontupper=\ttfamily\scriptsize]
<*> <*> NCCL INFO comm <*> rank <*> nranks 2 cudaDev <*> busId <*> - Abort COMPLETE
\end{tcolorbox}
    \vspace{-4mm}
\caption{Drain output for two NCCL log lines from different ranks.}
\label{fig:drain-template}
\end{figure}

\noindent\textbf{RQ3: Effectiveness--Cost Trade-off.}
Table~\ref{tab:cost-rawvsparsed-results} reports per-run averages. Relative to parsed input, raw input increases average input-token usage by $10.8\times$ for CoT and $9.0\times$ for ICL. Corresponding cost increases are $7.8\times$ and $8.9\times$. LLM latency rises from 1.961\,s to 2.176\,s for CoT and from 1.012\,s to 1.943\,s for ICL.

These measurements expose a direct effectiveness–efficiency trade-off. Drain-based parsing improves efficiency but can discard information that affects detection.

\begin{table}[]
\caption{LogPrompt cost: raw vs.\ parsed logs (per-run averages).}
    \vspace{-3mm}

\label{tab:cost-rawvsparsed-results}
\scriptsize
\setlength{\tabcolsep}{8pt}
\renewcommand{\arraystretch}{1.0}
\begin{tabular}{@{}lcccc@{}}
\toprule
\textbf{Variation} & \textbf{Input Tokens} & \textbf{Output Tokens} & \textbf{Cost (USD)} & \textbf{LLM Latency (s)} \\
\midrule
CoT \scriptsize{(Parsed)} & 1,282.5  & 92.8  & 0.0046 & 1.961 \\
CoT \scriptsize{(Raw)}    & 13,789.2  & 92.0  & 0.0359 & 2.176 \\
\cmidrule(lr){1-5}
ICL \scriptsize{(Parsed)}  & 4,617.5 & 8.0  & 0.0117 & 1.012 \\
ICL \scriptsize{(Raw)} & 41,396.2 & 8.0  & 0.1036 & 1.943 \\
\bottomrule
\end{tabular}
\end{table}

\section{Discussion}
\textbf{Threats to Validity.} Our findings are initial, in-distribution evidence for one configuration: two A40 GPUs, two injected fault classes, NCCL logs alone, Drain as the only parser, and a single labeled demonstration per class for ICL-LogPrompt, which the sequence models do not receive. Results may also depend on rank-ordered concatenation, demonstration choice, the GPT-5.4 version, and LLM nondeterminism, since no seed was fixed and each run was queried once. The parsing stage's own overhead and FPR under operational prevalence were not measured. Additional parsers, observability sources, fault mechanisms, and cross-environment replication are needed to establish generality.

\noindent\textbf{Conclusions.} We introduced \dataset, a PyTorch-based framework for injecting controlled faults and collecting observability data from multi-GPU workloads, and constructed an initial 220-run NCCL evaluation set. 
Our empirical evaluation shows that established log-anomaly detection methods transfer poorly to this setting: on Drain-parsed logs, the highest overall F1 is 0.667, and no method detects both failure classes reliably. Replacing Drain-parsed input with raw logs raises ICL-LogPrompt's F1 to 0.983, while increasing input-token usage by $9.0\times$ and per-run cost by $8.9\times$. This suggests Drain-based parsing can discard distributed execution context that detection depends on.
Future releases will add broader memory, runtime, communication, hardware, and application faults while retaining the current reproducible design.

\section*{Data Availability Statement}
The source code and experimental artifacts of this work are publicly available at \url{https://github.com/cuiboyuan/nccl_log_rca_bench}.

\balance
\bibliographystyle{ACM-Reference-Format}
\bibliography{reference}

\end{document}